\documentclass[sigconf]{acmart}

\usepackage{booktabs}
\usepackage{tabularx}
\usepackage{placeins}
\usepackage{amsmath}
\usepackage{listings}

\AtBeginDocument{%
  }

\setcopyright{acmlicensed}
\copyrightyear{2024}
\acmYear{2024}
\acmDOI{}

\acmConference[KDD Cup CRAG Workshop '24]{2024 KDD Cup Workshop for Retrieval Augmented Generation}{August 25--29,
  2024}{Barcelona, Spain}

\acmISBN{978-1-4503-XXXX-X/18/06}

\begin{document}

\title[WeKnow-RAG: An Adaptive Approach for Retrieval-Augmented Generation ...]{WeKnow-RAG: An Adaptive Approach for Retrieval-Augmented Generation Integrating Web Search and Knowledge Graphs}
\author{Weijian Xie}
\affiliation{%
  \institution{Ping An Bank Co., Ltd.}
  \city{Shenzhen}
  \state{Guangdong}
  \country{China}}
\email{tsewkviko@gmail.com}

\author{Xuefeng Liang}
\affiliation{%
  \institution{Tsinghua University}
  \city{Haidian}
  \state{Beijing}
  \country{China}}
\email{liang1xuefeng@gmail.com}

\author{Yuhui Liu}
\affiliation{%
  \institution{Central South University}
  \city{Changsha}
  \state{Hunan}
  \country{China}}
\email{liuyuhui@csu.edu.cn}

\author{Kaihua Ni}
\affiliation{%
 \institution{Shanghai Yiwo Information Technology Co., Ltd.}
 \city{Changning}
 \state{Shanghai}
 \country{China}}
\email{nikaihua008@gmail.com}

\author{Hong Cheng}
\affiliation{%
  \institution{Kwangwoon University, Department of Computer Engineering}
  \city{Nowon Gu}
  \state{Seoul}
  \country{South Korea}}
\email{kwchenghong@gmail.com}

\author{Zetian Hu}
\affiliation{%
  \institution{Tsinghua University}
  \city{Haidian}
  \state{Beijing}
  \country{China}}
\email{huzt22@mails.tsinghua.edu.cn}



\renewcommand{\shortauthors}{Xie et al.}

\begin{abstract}
Large Language Models (LLMs) have greatly contributed to the development of adaptive intelligent agents and are positioned as an important way to achieve Artificial General Intelligence (AGI). However, LLMs are prone to produce factually incorrect information and often produce "phantom" content that undermines their reliability, which poses a serious challenge for their deployment in real-world scenarios. Enhancing LLMs by combining external databases and information retrieval mechanisms is an effective path. To address the above challenges, we propose a new approach called \textbf{WeKnow-RAG}, which integrates \textbf{We}b search and \textbf{Know}ledge Graphs into a "Retrieval-Augmented Generation (\textbf{RAG})" system. First, the accuracy and reliability of LLM responses are improved by combining the structured representation of Knowledge Graphs with the flexibility of dense vector retrieval. WeKnow-RAG then utilizes domain-specific knowledge graphs to satisfy a variety of queries and domains, thereby improving performance on factual information and complex reasoning tasks by employing multi-stage web page retrieval techniques using both sparse and dense retrieval methods. Our approach effectively balances the efficiency and accuracy of information retrieval, thus improving the overall retrieval process. Finally, we also integrate a self-assessment mechanism for the LLM to evaluate the trustworthiness of the answers it generates. Our approach proves its outstanding effectiveness in a wide range of offline experiments and online submissions.
\end{abstract}




\maketitle

\section{Introduction}
\label{_Introduction}
Large language models (LLMs) \cite{rajpurkar2016squad} have significantly propelled the development of adaptable intelligent agents, positioning themselves as a promising pathway towards achieving artificial general intelligence (AGI). Despite these advancements, the inherent nature of LLMs introduces critical challenges that obstruct their deployment in real-world production scenarios \cite{ji2023survey}. One of main key issues is the tendency of LLMs to produce factually incorrect information, often resulting in "hallucinated" content that undermines their reliability \cite{zhang2024siren}. \cite{vu2023freshllms}  demonstrated that GPT-4's exhibits excellent performances in responding to questions about both slow-changing or fast-changing facts, with an accuracy rate below 15\%. However, even for stable (never-changing) facts, GPT-4's accuracy in addressing questions about less popular (torso-to-tail) entities is under 35\% \cite{sun2023head}.

Researchers have implemented a variety of methods to enhance knowledge in LLMs, one of which is Retrieval-Augmented Generation (RAG) method. RAG method enhances LLMs by incorporating external databases and information retrieval mechanisms \cite{yu2022survey}. This technique dynamically incorporates relevant information into the LLMs' prompt during inference without altering the weights of the used LLMs. By integrating external knowledge, RAG methods can reduce hallucinations of LLMs achieving better performance that can surpass traditional fine-tuning approaches, especially in applications that require high precision and up-to-date information \cite{lewis2020retrieval}.

Implementations of RAG methods typically depend on the dense vector similarity search for retrieval. Yet, this technique, which divides the corpus into text chunks and uses dense retrieval systems exclusively, falls short for complex queries \cite{lewis2020retrieval}. Some methods address this issue by employing metadata filtering or hybrid search techniques \cite{wu2022hqann}, but these methods are constrained by the pre-defined scope of metadata by developers. Furthermore, achieving the necessary granularity to answer complex queries within similar vector space chunks remains challenging \cite{kong2022multi}. The inefficiency comes from the method's inability to selectively retrieve relevant information, leading to large amounts of chunk data retrieval that may not directly answer the queries \cite{Luan2020Sparse}. 

An ideal RAG system should retrieve only the essential content, reducing the inclusion of irrelevant information. This is where Knowledge Graphs (KGs) can assist by providing a structured and explicit representation of entities and relationships that are more precise than information retrieval through vector similarity \cite{mohoney2023high}. KGs enable searching for "things, not strings" by maintaining extensive collections of explicit facts structured as accurate, mutable, and interpretable knowledge triples \cite{Cambria2021Knowledge}. A knowledge triple typically represents a fact in the format (entity) - relationship → (entity) \cite{dorpinghaus2020knowledge}. In addition, KGs can expand by continuously incorporating new information, and experts can build domain-specific KGs to deliver precise and trustworthy data within particular fields \cite{An2022Exploring}. Well-known KGs such as Freebase, Wikidata, and YAGO could serve as practical examples of this concept. Considerable researches have emerged at the crossroads of graph-based methodologies and LLMs, showcasing applications like reasoning over graphs and enhancing the integration of graph data with LLMs \cite{Choudhary2023Complex}.

To tackle the above challenges, we propose a novel approach named \textbf{WeKnow-RAG} that integrates \textbf{We}b Search and \textbf{Know}ledge Graphs in a Retrieval-Augmented Generation (\textbf{RAG}) system. Our approach combines the structured representation of KGs with the flexibility of dense vector retrieval to enhance the accuracy and reliability of LLM responses.

The main contributions of our work are as follows:

\begin{itemize}
    \item We develop a domain-specific KG-enhanced RAG system that adapts to different types of queries and domains, improving performance on both factual and complex reasoning tasks.
    \item We introduce a multi-stage retrieval method for web pages that leverages both sparse and dense retrieval techniques, effectively balancing efficiency and accuracy in information retrieval.
    \item We implement a self-assessment mechanism for LLMs to evaluate their confidence in generated answers, reducing hallucinations and improving overall response quality.
    \item We present an adaptive framework that intelligently combines KG-based and web-based RAG methods based on the characteristics of different domains.
\end{itemize}

The proposed WeKnow-RAG won 3rd place in the final evaluation of Task 3 at the Meta KDD CUP 2024, which was assessed using the Comprehensive RAG Benchmark (CRAG), a factual question-answering benchmark consisting of numerous question-answer pairs and mock APIs to simulate web and Knowledge Graph search. Our experimental results on the CRAG dataset demonstrate the effectiveness of our approach, achieving significant improvements in accuracy and reducing hallucinations across various domains and question types.

\section{Related Work}\label{_RelatedWork}
In order to improve the accuracy and reliability of LLMs in question answer tasks, many previous works have proven effective. According to whether the parameters of LLMs to be modified or not, we categorize them into two main approaches \cite{huang2023surveyhallucinationlargelanguage}:
\begin{itemize}
    \item \textbf{Fine-tuning and calibration}: Fine-tuning the LLM on a specific domain or task can improve its accuracy and reduce hallucinatory responses. In addition, calibrating the model to provide uncertainty estimates with responses can help users assess the reliability of the generated information\cite{zhang2023instruction}. This approach requires modifying the parameters of the LLMs.
    \item \textbf{External Knowledge Integration}: Integrating external knowledge sources into LLM can help enhance its comprehension and reduce hallucinatory responses.
    These sources can be massively updated web pages, databases, knowledge graphs, additional LLMs, etc. RAG is a method of utilizing many knowledge sources to provide informed answers, thereby improving the accuracy of LLMs' generated responses\cite{gao2024retrievalaugmented}. This approach does not require modification of the parameters of the LLMs. 
\end{itemize}

\subsection{Fine-tuning and calibration}
High-quality QA datasets are crucial for LLMs fine-tuning and calibration. These datasets typically contain a diverse and well-structured collection of questions and answers, enabling the model to learn and generalize across different scenarios.There are some famous high-quality QA dataset:
\begin{itemize}
  \item \textbf{SQuAD}\cite{rajpurkar2016squad}:Consisting of over 100,000 questions posed on a set of Wikipedia articles. The answers are often spans of text directly extracted from the given passages.
  \item \textbf{Natural Questions}\cite{kwiatkowski2019natural}:Created by Google, this dataset contains real questions from Google Search users, with answers provided as spans of text from Wikipedia pages.
  \item \textbf{TriviaQA}\cite{joshi2017triviaqa}:Contains over 650,000 question-answer pairs, where the questions are sourced from trivia websites, and the answers are verified against evidence documents.
  \item \textbf{MS MARCO}\cite{bajaj2018msmarco}:A large-scale dataset that includes real anonymized queries from Bing search logs, with answers generated by humans based on a set of retrieved documents.
\end{itemize}

However, in practical QA scenarios like this competition that contain a lot of noisy information and html code, it's hard to use this dataset directly for fine-tuning. And these open source datasets only contains a small amount of QA-pairs in special domain like finance, sports, music and movie.
\subsection{External Knowledge Integration}
Recent endeavors have leveraged RAG to improve LLMs across diverse tasks \cite{borgeaud2022improving,izacard2020distilling,weijia2023replug}, particularly those requiring up-to-date and accurate knowledge such as question answering (QA), AI4Science, and software engineering. For instance, Lozano et al. \cite{lozano2023clinfo} developed a scientific QA system that dynamically retrieves scientific literature. MolReGPT \cite{li2024empowering} uses RAG to boost ChatGPT's in-context learning for molecular discovery. Moreover, RAG has been shown to effectively mitigate hallucinations in conversational tasks \cite{shuster2021retrieval,jing2021beyond}.

Recent progress in integrating LLMs with KGs has also been remarkable. Jin et al. \cite{jin2023large} reviewed this integration comprehensively, classifying LLM roles as Predictors, Encoders, and Aligners. For graph-based reasoning, Think-on-Graph \cite{sun2023think} and Reasoning-on-Graph \cite{luo2023reasoning} improve LLMs' reasoning capabilities by incorporating KGs. Yang et al. \cite{yang2023chatgpt} suggest enhancing LLMs' factual reasoning in different training stages using KGs. In LLM-based QA, Wen et al.'s Mindmap \cite{wen2023mindmap} and Qi et al. \cite{qi2023foodgpt} utilize KGs to advance LLM inference in specific fields like medicine and food. These developments illustrate the growing effectiveness of combining LLMs and KGs in improving information retrieval and reasoning.

\section{Preliminaries}\label{_PRELIMINARIES}

A RAG QA system takes a question $Q$ as input and outputs an answer $A$; the answer is generated by LLMs according to information retrieved from external sources or directly from the knowledge internalized in the model \cite{yang2024cragcomprehensiverag}. The answer $A$ should provide useful information to answer the question without adding any hallucinations under ideal conditions. 

\textbf{Task description}
The process to obtain an answer from a question can be called a query. For each question $Q$, both web search results and Mock APIs for information retrieval are available for the generation of answer $A$. The web search results comprise 50 web page candidates. The Mock Knowledge Graphs (MKGs) contain structured data pertinent to the queries; the answers may or may not be present in the MKGs. The Mock APIs accept input parameters, which are often derived from a query, and deliver structured data from the MKGs to assist in answer generation. 

RAG systems are evaluated using a scoring method that rates response quality as correct (1 point), missing (0 points), and incorrect (-1 point). A response is rated as missing if it is 'I don't know'. Otherwise, an LLM will be used to determine whether the response is correct or incorrect \cite{yang2024cragcomprehensiverag}.

\textbf{Dataset description}
The dataset used is the Comprehensive RAG Benchmark, covering five domains: finance, sports, music, movie, and open domain, and eight types of questions as detailed in \cite{yang2024cragcomprehensiverag}. Specifically, CRAG includes questions with answers ranging from seconds to years; it considers the popularity of entities and covers not only head but also torso and tail facts. It contains simple factual questions as well as seven types of complex questions, including comparison, aggregation, set questions and so on, to test the reasoning and synthesis capabilities of RAG solutions.

For the web search results, we used the question text as the search query and stored up to 50 web pages retrieved from the Brave search API. Each web page contains some key-value pairs, detailed in the following data schema:
\begin{itemize}
\item {Page Name}: The name of the webpage.
\item {Page Result}: The full HTML of the webpage.
\item {Page Snippet}: A short paragraph describing the major content of the page.
\end{itemize}

To construct the MKGs, we employed the CRAG KG APIs supplied by the organizer. These APIs incorporated publicly accessible KG data as well as randomly chosen entities of the same type and 'hard negative' entities with similar characteristics. The CRAG KG APIs were formulated with specific parameters to enable structured searches within the MKGs, resulting in a compilation of 38 mock APIs \cite{yang2024cragcomprehensiverag}.


\FloatBarrier
\begin{figure*}[!htbp]
  \centering
  \includegraphics[width=0.8\linewidth]{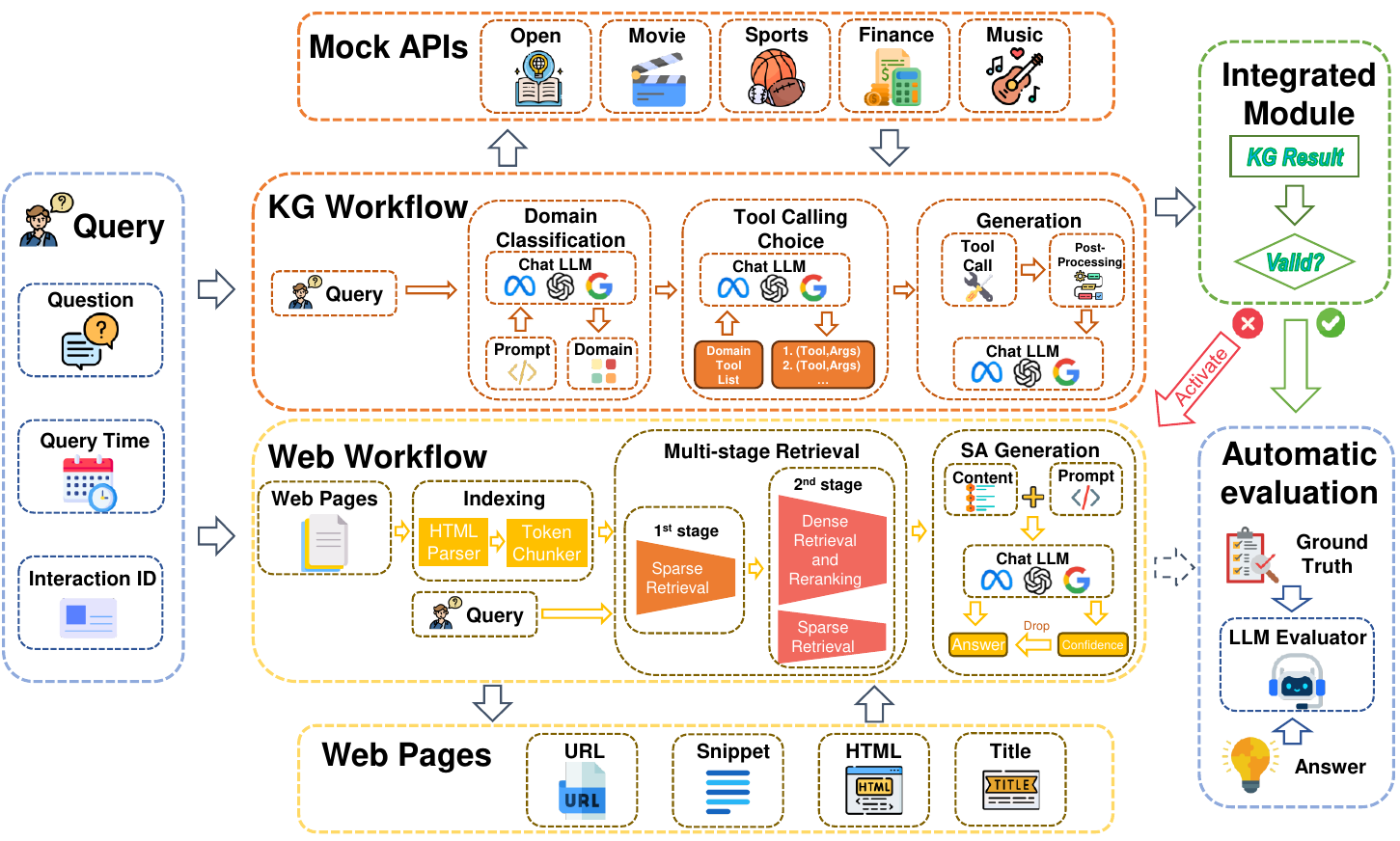}
  \caption{WeKnow-RAG pipeline for End-to-End Retrieval-Augmented Generation.}
  \label{fig:framework}
\end{figure*}

\section{Methods}
\label{_Method}
We proposed a approach named WeKnow-RAG for End-to-End Retrieval-Augmented Generation, as illustrated in Fig. \ref{fig:framework}. The pipeline comprises a KG workflow and a web search workflow to address the End-to-End Retrieval-Augmented Generation challenge, ultimately integrating both effectively. We will detail each component of the pipeline in the following sections.

\subsection{Web-based RAG}
\subsubsection{Web Content Parsing}
To utilize the data source in RAG approaches, content parsing is a critical process. It helps to comprehend unstructured data, convert it into structured data, and acquire the necessary information to answer the question.


To obtain more complete content for the source page, we parse the original HTML source code with the BeautifulSoup library\footnote{BeautifulSoup: \url{https://www.crummy.com/software/BeautifulSoup/}}.

\subsubsection{Chunking}
Chunking is the process of dividing a document into multiple paragraphs. The effectiveness of chunking directly influences the performance of question-answering systems.

Various chunking strategies exist, including sentence-level, token-level, and semantic-level approaches \cite{wang2024searching}. In our methods, we opt for token-level chunking for ease of use.

To determine the optimal configurations, a series of experiments are carried out to compare various chunk sizes in Section \ref{_Ablation}.

\begin{figure}[!htbp]
  \centering
  \includegraphics[width=\linewidth]{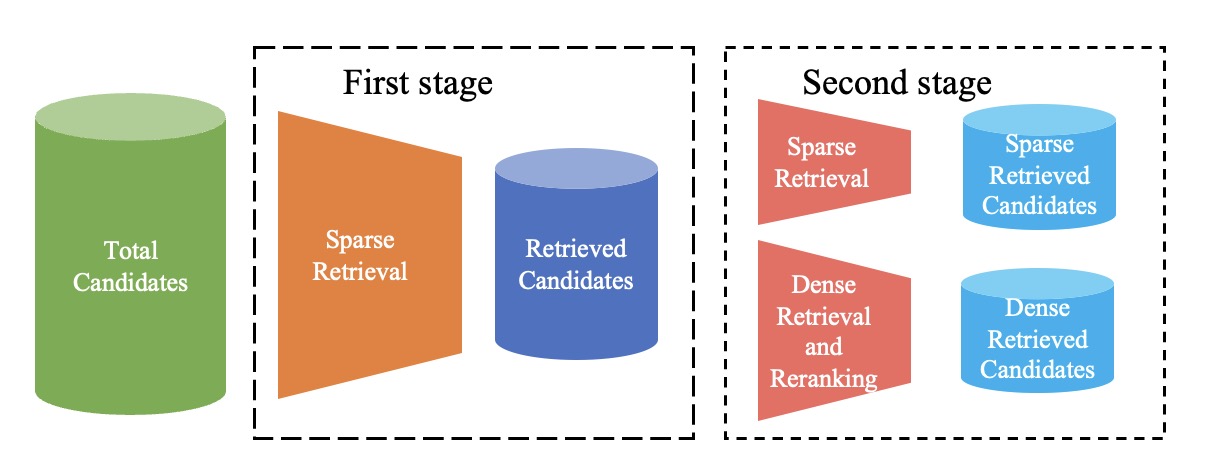}
  \caption{Multi-stage Retrieval methods.}
  \Description{The retrieval process consists of two stages: sparse retrieval in the first stage and hybrid search in the second stage.}
  \label{fig:retrieval}
\end{figure}

\subsubsection{Multi-stage Retrieval}

In the context of RAG, efficiently retrieving relevant documents from the data source is crucial for obtaining accurate answers and minimizing hallucinations.

We employ a multi-stage retrieval approach, as illustrated in Figure \ref{fig:retrieval}, to strike a balance between the effectiveness and efficiency of retrieving relevant paragraphs based on a query.

The explanation provided in Figure \ref{fig:retrieval} effectively outlines the initial stage of utilizing sparse retrieval to gather candidates from page result chunks and snippet chunks. It clarifies that sparse retrieval is employed to efficiently obtain pertinent paragraphs. Additionally, it highlights the selection process of the top $K$ candidates based on their BM25 scores, as denoted in Eq. \ref{eq_BM25}.
\begin{equation}\label{eq_BM25}
\begin{aligned}
& \text{Score}(query, C_i) = \\
& \sum_{q_j \in query}\text{IDF}(q_j) \cdot \frac{f(q_j,\text{C}_i) \cdot (k_1 + 1)}{f(q_j, \text{C}_i) + k_1 \cdot (1 - b + b \cdot \frac{|\text{C}_i|}{\text{avg\_dl}})}.
\end{aligned}
\end{equation}
Where:
\begin{itemize}
\item $q_j$ is a term in the query. $\text{IDF}(q_j)$ is the inverse document frequency of term  $q_j$.
\item $ f(q_j, \text{C}_i) $ is the term frequency of term  $q_j$ in document $ \text{C}_i $.
\item $ k_1 $ is a term frequency saturation parameter (typically set to 1.5).
\item $ b $ is a length normalization parameter (typically set to 0.75).
\item $ |\text{C}_i| $is the length of document $ \text{C}_i$.
\item $ \text{avg\_dl} $ is the average document length in the corpus.
\end{itemize}



By leveraging the strengths of different algorithms, hybrid retrieval can achieve better performance than any single algorithm. In the second stage, We employ hybrid search using the most common pattern \cite{gao2024retrievalaugmented}, which combines a sparse retriever (BM25) with a dense retriever (embedding similarity), as their strengths are complementary. The sparse retriever excels at finding relevant documents based on keywords, while the dense retriever identifies relevant documents based on semantic similarity.

The sparse retrieval approach in the second stage uses the BM25 score, just like in the first stage, to select the top $M$ candidates based on the highest score.

The dense retrieval approach in second stage comprises dense embedding retrieval and reranking methods. A bge-large-en-v1.5 \cite{bge_embedding} embedding model is employed for dense embedding retrieval, selecting the top $N$ candidates based on the highest embedding similarity with the query. Subsequently, bge-reranker-large \cite{bge_embedding} reranker model is applied to obtain a more accurate relevance score and rerank the top $N$ candidates in the reranking approach of Figure \ref{fig:retrieval}.

Notice that $K >> M + N$, which allowed us to obtain K relevant candidates in the first stage with low latency. This enabled us to conduct dense retrieval to obtain N relevant candidates and reranking using models with more parameters (such as the bge-reranker-large model, which has 0.5B parameters) to obtain M relevant candidates with fewer data in the second stage.

It is worth mentioning that to include as much relevant information as possible, we added not only the parsed HTML content in the database but also the page name and page snippet for retrieval.

\subsubsection{Answer Generation with Self-Assessment}
\label{_AnswerGeneration}
Hallucination is a common issue in the generated content of LLMs. In order to address this challenge, we propose a self-assessment mechanism that evaluates the confidence level of the generated answer and determines its suitability for selection.

In particular, we instruct the LLM to indicate the confidence level (high, medium, low) corresponding to the generated answer. We decide to accept the answer only when the confidence level meets the specified requirement. If the confidence level is below the threshold, we conclude that the LLM lacks sufficient confidence to answer the question and will output "I don't know". Experiments on performance with various confidence level thresholds are detailed in Section \ref{_Ablation}, and the complete instructions for the LLM are provided in Appendix \ref{_AGPrompt}.


\subsection{Knowledge Graphs}
By extracting knowledge from structured and unstructured data sources, a domain-specific knowledge graph can be constructed. This includes entities, relationships, and facts related to each domain while using state-of-the-art information extraction techniques to ensure that the knowledge is rich and up-to-date. This type of knowledge base plays a key role in various question-answering tasks, drawing inspiration from the construction of large knowledge bases such as DBpedia\cite{auer2007dbpedia}. By parsing a question into step-by-step sub-questions or sub-functions, establishing the relationship between the sub-questions and the model application programming interface (API), and solving a series of sub-questions using conditional lookups, set searches, comparisons, aggregations, multi-hop queries, and post-processing operations, the entire question can be answered.

\subsubsection{Domain classification}
Our method categorizes the question into domains by making an initial call to an LLM. At this point, the LLM acts as an intelligent linguistic categorization tool. For example, for the question, "How much did Funko open at today?", it is automatically categorized as the financial domain. For the question, "What label is Taylor Swift signed to?", it is automatically categorized as the music domain. Domain categorization is critical for subsequent processing. For this reason, in the four domains of movie, sport, finance, and music, we suggest that the model should only make a determination if it has more than 90\% certainty. When certainty is below this threshold, the domains are categorized as open domains.

\subsubsection{Query generation}
\label{_QueryGeneration}
Depending on the domain categorization of the problem, the corresponding domain cueing phase is entered, where the second call to the LLM is made. The hints provided are different for each domain, and the model needs to return structured analysis results. For example, in the music domain, it needs to return three kinds of functions, categorized into artist, song, and year, with each category containing a variety of attribute information. For the instruction to generate the function calling query, refer to Appendix \ref{_QGPrompt}. Simultaneously, questions that do not fall into these three categories are answered directly using the model or the RAG workflow. The query generator will transform the analysis results into structured queries compatible with the KG API. For example, a multi-hop problem will be decomposed into a series of API calls, each of which resolves a link in the information chain.

\subsubsection{Answer Retrieval and Post-processing}
A set of candidate answers is retrieved by performing a structured query on the KG through an application programming interface (API). For post-processing problems, additional reasoning is applied. We employ a rule-based system that utilizes machine learning techniques to handle temporal reasoning, numerical computation, and logical reasoning \cite{chen2020hybridqa}. With this hybrid approach, we can identify inconsistencies between the retrieved data and the problem hypotheses, effectively solving challenging false premise problems.

\subsection{Integrated method}
The diverse nature of the CRAG dataset requires a dynamic approach that adapts to the distinct characteristics of each domain, particularly in terms of the velocity at which query information evolves. To address this, we characterize the time distribution of each domain, by analyzing the key "static-or-dynamic" defined in the dataset, which are categorized as "static", "slow-changing", "fast-changing" and "real-time", and then propose an adaptive framework that intelligently balances the use of Knowledge Graphs (KGs) and Web-based RAG methods.

For stable domains such as the Encyclopedia Open domain, where the velocity of query information change is minimal, our system follows the rule of prioritizing the output of KG Workflow and not activating the whole Web-based RAG Workflow, as demonstrated in Figure \ref{fig:framework}. The robustness and reliability of KGs in these domains ensure high accuracy for questions that do not necessitate up-to-the-minute data. This approach aligns with the findings of Neumaier et al.\cite{neumaier2016automated}, who underscore the effectiveness of KGs in stable informational contexts.

For domains with gradual information change, such as Music and Movies, we maintain the primacy of KGs while incorporating periodic updates to capture the latest information. This strategy ensures that our KGs remain relevant for answering queries that may involve recent but not instantaneous changes. The update frequency is determined by a domain-specific change detection algorithm, which is controlled the LLM.

\section{Experiments}\label{_Experiments}

\subsection{Experimental settings}
\label{_Experiments_Settings}
We thoroughly introduce our experimental settings, encompassing the datasets, evaluation metrics, and parameter settings.

\textbf{Datasets.}
The entire CRAG dataset contains over 2000 questions, and testing on the whole dataset would be very time-consuming. During offline testing, to quickly iterate and optimize, we tested on a subset of the CRAG dataset containing 200 questions. 

\textbf{Evaluation metrics.}
For each question, we use a three-way scoring system, assigning 1 for correct answers, -1 for incorrect answers, and 0 for missing answers. An answer is considered accurate if it exactly matches the ground truth and missing if it is 'I don't know'. Otherwise, we use model-based automatic evaluation with GPT-4 to determine whether the response is correct or incorrect. There are four evaluation metrics: \textbf{Accuracy}, \textbf{Hallucination}, \textbf{Missing}, and \textbf{Score}. Accuracy, Hallucination, and Missing represent the percentage of accurate, incorrect, and missing answers in the test set, respectively, while Score is the difference between Accuracy and Hallucination.

\textbf{Parameter settings.}
We set the number $K$ of retrieving relevant candidates in the first stage of Web-based RAG as 200 for parsed page result chunks and page snippet chunks, respectively. The number $M$ of sparse retrieval candidates is set as 5, and the number of dense retrieval candidates $N$ is set as 20 in the second stage. $K$, $M$, $N$ are selected empirically and include correct answer information whenever possible. We employed the bge-large-en-v1.5 \cite{bge_embedding} as the dense retrieval model, bge-reranker-large \cite{bge_embedding} as the reranker model, and llama-3-70b-instruct-awq (an awq quantization version of Meta-Llama-3.1-70B-Instruct) as our Chat LLM model. All of them can be downloaded from the Hugging Face Hub\footnote{Hugging face: 
\url{https://huggingface.co/}}.





\subsection{Main results}
The final submission for Task 3 included two versions demonstrating online automated assessment performance, as shown in the Table \ref{tab:online}. Version 2 made three main improvements relative to Version 1. First, it enhanced the handling of prompts for open domains. Second, it improved the model's focus on false premise queries by including examples of such queries in the prompts. Third, the hints related to opening, closing, and price were adjusted. Overall, the strategy of the entire method involves first using the KG process to directly find answer-related information, strictly requiring the KG part to answer only questions with certainty to significantly reduce the error rate. The RAG process is then used to synthesize the information provided by the KG or from web pages to deliver the final answer to the question.

\begin{table}
\centering
\caption{Online submission results of our method on Task 3 in KDD Cup 2024 - CRAG Round 2.}
\begin{tabular}{lcccc}
\toprule
\textbf{Model} & \textbf{Accuracy} & \textbf{Hallucination} & \textbf{Missing} & \textbf{Score} \\
\midrule
version1 & 0.393 & 0.319 & 0.288 & 0.0743 \\
version2 & 0.409 & 0.316 & 0.276 & 0.0929 \\
\bottomrule
\end{tabular}
\label{tab:online}
\end{table}

The competition accepted a limited number of submissions. To thoroughly validate the method's enhancement, a local test set comprising 200 data points was designed offline, primarily focusing on validating the efficacy of the KGs component. The method's efficacy is underscored in Table \ref{tab:offline}. Initially, a classification prompt was devised for accurate categorization by the LLM, followed by a domain-specific prompt for movies, resulting in a base score of 0.064. Progressing to four domains and refining the strategy led to consistent improvement. While the KGs for individual domains were well-crafted, a lack of robust classification capability posed a significant constraint on further enhancement. To address this, an illegal problem optimization was integrated into the classification prompt, ensuring the model's classification into the sports, movies, music, and finance domains with a reasonable degree of certainty, boosting the score to 0.1499. Subsequent refinements focusing on detailed optimization for the finance domain and open domains further enhanced the performance.

\begin{table}
\centering
\caption{Experimental results of offline testing of our method.}
\begin{tabular}{lcccc}
\toprule
\textbf{Model} & \textbf{Acc.} & \textbf{Hall.} & \textbf{Missing} & \textbf{Score} \\
\midrule
Single domain & 0.100 & 0.035 & 0.865 & 0.0640 \\
+ Four domains & 0.125 & 0.025 & 0.850 & 0.1000 \\
+ Classification & 0.340 & 0.190 & 0.470 & 0.1499 \\
+ Open Optimize & 0.340 & 0.185 & 0.475 & 0.1550 \\
\bottomrule
\end{tabular} 
\label{tab:offline}
\end{table}


%

\subsection{Ablation study}
\label{_Ablation}
To enhance performance in each phase, a series of experiments are carried out using a subset of the CRAG dataset.

\textbf{Chunk size.} In the chunking process of RAG, the size of the chunk plays a critical role in obtaining semantically complete paragraphs. We conducted experiments with various chunk sizes as illustrated in Table \ref{tab:chunksize}. The results indicate that a chunk size of 750 yields the best performance in a small subset of CRAG dataset. In Round 1 submissions, we experimented with chunk sizes of 500 and 750. After comparing the results, we determined that a chunk size of 500 yielded better outcomes. Therefore, we have decided to stick with a chunk size of 500 for the final submission.
\begin{table}[h]
\centering
\caption{An analysis of various chunk sizes on a subset of the CRAG dataset using the Web-based RAG approach exclusively.}
\begin{tabular}{lcccc}
\toprule
\textbf{Chunk size} & \textbf{Accuracy}  & \textbf{Hallucination} & \textbf{Missing}  & \textbf{Score} \\ 
\midrule
300 & 0.38 & 0.35 & 0.27 & 0.03 \\ 
500 & 0.38 & 0.28 & 0.34 & 0.10 \\
750 & 0.41 & 0.28 & 0.31 & 0.13 \\  
1000 & 0.40 & 0.30 & 0.30 & 0.10 \\        
\bottomrule
\end{tabular}
\label{tab:chunksize}
\end{table}

\textbf{Confidence level.} In the self-assessment phase of answer generation, we conduct experiments with different confidence levels indicated by the LLM. The results in Table \ref{tab:confidence} show that when the confidence level is "high", it achieves the highest score. We found that by using the confidence level threshold, the extra accuracy is significantly enhanced.
\begin{table*}
\centering
\caption{An analysis of various confidence level thresholds on a subset of the CRAG dataset using the Web-based RAG approach exclusively.}
\begin{tabular}{lccccc}
\toprule
\textbf{Confidence level threshold} & \textbf{Exact Accuracy} & \textbf{Accuracy}  & \textbf{Hallucination} & \textbf{Missing}  & \textbf{Score} \\ 
\midrule
low (w/o Self-Assessment) & 0.05 & 0.38 & 0.28 & \textbf{0.34} & 0.10 \\
medium & \textbf{0.15} & 0.33 & 0.28 & 0.39 & 0.05 \\  
\textbf{high} & 0.14 & \textbf{0.38} & \textbf{0.26} & 0.36 & \textbf{0.12} \\        
\bottomrule
\end{tabular}
\label{tab:confidence}
\end{table*}



\subsection{Model analysis}

The experimental results above demonstrate that the WeKnow-RAG approach we proposed achieves effective performance by leveraging the contribution of each module in the Web-based RAG workflow and KG workflow.

KG workflow provides accurate answers with minimal errors by using function calling to extract specific information from knowledge graphs.

While the web-based RAG workflow provides more relevant information from a large number of web pages through multi-stage retrieval, it also reduces hallucination through a self-assessment approach.




\section{Conclusions}\label{_Conclusions}

While LLM has shown promising applications across various domains, highlighting substantial development opportunities and research significance, its intrinsic traits can lead to outputs that deviate from factual accuracy, generating "hallucinatory" content that poses a significant obstacle to research advancement.
To address the above challenges, we propose a new approach called WeKnow-RAG that integrates Web search and Knowledge Graphs into a RAG system. Our approach combines the structured representation of knowledge graphs with the flexibility of dense vector retrieval to improve the accuracy and reliability of LLM responses. By designing a specialized RAG system, WeKnow-RAG utilizes domain-specific knowledge graphs to satisfy a wide range of queries and domains, improving the performance of factual information and complex reasoning tasks. By employing multi-stage web page retrieval techniques, both sparse and dense retrieval methods are used. Our approach effectively balances the efficiency and accuracy of information retrieval, thus enhancing the entire retrieval process. Meanwhile, we integrate a self-assessment mechanism for LLMs to evaluate the trustworthiness of the answers they generate. This mechanism reduces illusions and thus improves the quality of model-generated answers.The WeKnow-RAG framework introduces an adaptive methodology that intelligently combines KG-based RAG methods with web-based RAG methods. This integration is tailored to the unique characteristics of different domains and adapts to the speed of information evolution, thus ensuring optimal performance in dynamic information environments. Our approach has demonstrated outstanding effectiveness in extensive offline experiments and online submissions.

\section{Acknowledgments}
The competition was successfully concluded with thanks to the scientists and engineers from Meta Reality-Labs and the Hong Kong University of Science and Technology (HKUST, HKUST-GZ) for their great support and valuable data during the competition. The authors also acknowledge the computer resources given by Central South University’s High Performance Computing Center (HPC).

\bibliographystyle{ACM-Reference-Format}
\bibliography{sample-base}

\appendix
\section{Prompts}
\subsection{Answer Generation with Self-Assessment}
\label{_AGPrompt}

The prompt for Web-based RAG to generate an answer and the corresponding confidence level is displayed below.

\begin{lstlisting}
(*@\textbf{System:}@*)
You are provided with a question, current time and various references. Your task is to answer the question succinctly, using the FEWEST words possible. If you are absolutely sure, more than 97\% confident, please answer directly. If you are not sure, please respond with 'I don't know'.
Please answer the question and provide the confidence tier (high, medium, low) for your answer. Use the following standards for confidence tiers:

- High Confidence (High): The answer provided is almost certainly correct. There is strong evidence or overwhelming consensus supporting this answer. The model has a high level of certainty and little to no doubt about this answer.
- Medium Confidence (Medium): The answer provided is likely to be correct. There is some evidence or reasonable support for this answer, but it is not conclusive. The model has some level of certainty but acknowledges that there is a possibility of error or alternative answers.
- Low Confidence (Low): The answer provided is uncertain or speculative. There is little to no solid evidence or support for this answer. The model has significant doubts about the accuracy of this answer and recognizes that it could easily be incorrect.

Output the result in JSON format, answer is a string, confidence is a string (value is one of high, medium, low), for example output: {"answer": "mayon volcano", "confidence": "medium"}

(*@\textbf{User:}@*)
Context: {references}
Current Time: {query_time}
Question: {query}
Output:
\end{lstlisting}

\subsection{Query Generation for KG function calling}
\label{_QGPrompt}

The prompt for generating function calls for querying knowledge graphs is presented below. It displays a part of the function call prompt in the Music Domain, the complete version includes several additional functions.

\begin{lstlisting}
(*@\textbf{System:}@*)
You are an AI agent of linguist talking to a human. ... For all questions you MUST use one of the functions provided.
You have access to the following tools:
{
    "type": "function",
    "function": {
        "name": "get_artist_info",
        "description": "Useful for when you need to get information about an artist, such as singer, band",
        "parameters": {
            "type": "object",
            "properties": {
                "artist_name": {
                    "type": "string",
                    "description": "the name of artist or band",
                },
                "artist_information": {
                    "type": "string",
                    "description": "the kind of artist information, such as birthplace, birthday, lifespan, all_works, grammy_count, grammy_year, band_members",
                }
            },
            "required": ["artist_name", "artist_information"],
        },
    },
}
...
To use these tools you must always respond in a Python function call based on the above provided function definition of the tool!
For example:
{"name": "get_artist_info", "params": {"artist_name": "justin bieber", "artist_information": "birthday"}}

(*@\textbf{User:}@*)
{query}
\end{lstlisting}








\end{document}